\documentclass[3p, times,procedia]{elsarticle}

\usepackage{ecrc}
\usepackage{amsmath}
\usepackage{amssymb}
\usepackage{booktabs}
\usepackage{float}
\usepackage{tikz}
\usetikzlibrary{arrows.meta,positioning}
\usepackage[bookmarks=false]{hyperref}
\hypersetup{colorlinks=true,linkcolor=blue,citecolor=blue,urlcolor=blue}

\newif\ifcameraready
\ifdefined\cameraReadyVersion
  \camerareadytrue
\else
  \camerareadyfalse
\fi

\makeatletter
\def\ps@pprintTitle{%
\vspace*{1.6mm}
    \def\@evenhead{%
      \setbox1=\hbox{\elslogo}%
      \setbox2=\hbox{\sdlogo}%
      \setbox3=\hbox{\jnllogo}%
      \vspace*{2pc}%
      \parbox[t]{\wd1}{\hspace*{-2.6pt}\raisebox{3pt}{\elslogo}}%
       \hfil\parbox[t]{19pc}{\centering%
       \raisebox{33.6pt}{\sdlogo}\\[-15pt]
       \mbox{\footnotesize\@journalname~00~(2026)~000--000}}\hfil%
        \raisebox{24.8pt}{\parbox[c]{\wd3}{\jnllogo\\[2pt]
        \ifelsarticle@nsmodel
         \hspace*{-2pc}{\footnotesize www.elsevier.com/locate/procedia}%
        \fi
    }}}
      \let\@oddhead\@evenhead
      \def\@oddfoot{\smash{\lower-19.3pt\vbox to 0pt{\fontsize{8}{10}\selectfont{\@issn~\@CopyrightLineBottom.\endgraf\hfill}}}}
}
\def\ps@headings{%
    \def\@oddhead{{\itshape\fontsize{8}{8}\selectfont
         \hfil\@runauth~/~\@journalname~00~(2026)~000--000\hfil}{\fontsize{8}{8}\selectfont\thepage}}
    \def\@evenhead{{\fontsize{8}{8}\selectfont\thepage}{\hfil{\itshape\fontsize{8}{8}\selectfont
         \@runauth~/~\@journalname~00~(2026)~000--000}}\hfil}
    \let\@evenfoot\@empty
    \let\@evenfoot\@oddfoot}
\gdef\@copyrightyear{2026}
\gdef\@copyrighttext{The Authors. Published by Elsevier B.V.\newline
This is an open access article under the CC BY-NC-ND license
(\href{http://creativecommons.org/licenses/by-nc-nd/4.0/}{http://creativecommons.org/licenses/by-nc-nd/4.0/})\newline
Peer-review under responsibility of the Conference Program Chairs}
\gdef\@CopyrightLine{\par\vskip1pc\noindent\textcopyright~2026~\@copyrighttext~}
\gdef\@CopyrightLineBottom{\noindent\textcopyright~2026~\@copyrighttext}
\makeatother
\volume{00}
\firstpage{1}
\journalname{Procedia Computer Science}
\ifcameraready
  \runauth{\cameraReadyRunningAuthors}
\else
  \runauth{Nora Girda and Adrian Groza }
\fi
\jid{procs}
\biboptions{numbers,sort&compress}

\begin{document}
\begin{frontmatter}

\dochead{The 13th International Workshop on Privacy and Security in Healthcare (PSCare 2026)\\October 28--30, 2026, Almaty, Kazakhstan}

\title{Review Before Trust: Source-Grounded Integrity Gates for AI-Assisted Personal Health Records}

\ifcameraready
  \author[utcn]{\cameraReadyAuthorOne}
  \author[utcn]{\cameraReadyAuthorTwo}
  \address[utcn]{\cameraReadyAffiliation}
\else
  \author{Nora Girda and Adrian Groza \\ Artificial Intelligence Research Institute AIRi@UTCN \\ Technical University of Cluj-Napoca, Cluj-Napoca, Romania \\
  nora.girda@campus.utcluj.ro, adrian.groza@cs.utcluj.ro}
\fi

\begin{abstract}
Large language models can convert medical documents into structured data, but plausible output may still be unsupported by the source. Persisting such output in a longitudinal health record, a record that accumulates patient information over time, therefore creates an integrity risk: unverified data may influence later summaries, trends, or preventive-care computations. We introduce an evidence-gated trust-promotion model that keeps generated data provisional until a deterministic monitor verifies it against the source document. The monitor admits a candidate for a specified downstream use only when the source contains a unique supporting quotation, the relevant fields occur within the same laboratory row, and the required provenance is preserved. The generator cannot approve its own output, missing or ambiguous evidence causes refusal, and refused candidates remain available for human review rather than being silently discarded. We implement the model in Medical DataCloud, a personal health-record application, and evaluate it through automated tests and a replay of saved extraction outputs. All 22 conformance and mutation tests pass. The replay covers nine historical laboratory PDF reports containing 102 manually labelled rows. The reports produce 97 numeric candidates: schema validation accepts all 97, an earlier packet-level evidence check accepts 94, and the hardened quotation- and row-level policy admits 72 while retaining 25 for review. The study evaluates system integrity rather than clinical correctness or clinical safety. The results demonstrate the technical feasibility of an enforceable boundary that prevents generated claims from authorizing their own reuse in a longitudinal health record.
\end{abstract}

\begin{keyword}
health data integrity \sep large language models \sep reference monitor \sep provenance \sep personal health records \sep abstention
\end{keyword}

\end{frontmatter}

\section{Introduction}

Large language models (LLMs) can extract structured information from clinical text with limited task-specific adaptation~\cite{agrawal2022,hein2025}. Extraction, however, is only the first step in a persistent health-data workflow. A generated laboratory value may later appear in a longitudinal trend, a preventive-care rule, an export, or the context supplied to another model. Persistence therefore turns a transient model output into a reusable system input. The resulting integrity question is not only whether the value appears plausible, but whether the system should permit a particular downstream use.

We use \emph{personal health record} to mean a longitudinal workspace in which a person stores medical documents and structured observations for later review and reuse. Within such a workspace, an LLM produces \emph{candidates}: schema-conforming propositions extracted from a source, such as an analyte, numeric value, unit, and reference interval. A candidate is not yet an authoritative observation. We call the unsupported transition from candidate to reusable record data \emph{unauthorized promotion}. For example, a schema-valid value supported only by a confidence score could silently enter a trend even when the cited row contains another number.

Neither confidence nor citation alone resolves this problem. Confidence remains generator-controlled or model-dependent, while a citation may be absent, ambiguous, duplicated, or unrelated to the asserted value~\cite{wu2025}. Human review can detect errors, but review guidance alone does not prevent an unchecked write. Governance frameworks similarly motivate oversight and traceability, yet an application still needs an operative rule that decides which generated claims may enter protected record functions~\cite{who2024,akgun2026,nist2024}.

This paper treats every generated candidate as untrusted input. A separate deterministic gate compares the candidate with preserved source text and decides whether the candidate may be projected into an observation for a specified purpose. In this paper, \emph{trusted} means admitted under that explicit integrity policy. It does not mean clinically correct, authentic, or safe. A refused candidate remains available as an extracted fact marked for review, but it does not enter the observation history used by downstream functions.

Medical DataCloud provides the concrete setting for this study. The application preserves uploaded documents, extracts bounded page and line packets, requests schema-constrained medical candidates from a hosted LLM, verifies source evidence in server code, stores both verified and refused candidates as source-linked facts, and projects only verified numeric laboratory candidates into observations. Those observations support longitudinal views, preventive workflows, and export. The laboratory path therefore exposes a security-sensitive transition that can be inspected and tested end to end.

The paper makes three contributions. First, it defines a purpose-scoped transition model with five integrity properties: mediated promotion, non-self-certification, provenance preservation, fail-closed admission, and uncertainty preservation. Second, it identifies the threat model and trusted computing base for a narrow laboratory path, then maps the model to independent evidence validation, verified-only projection, and a repository-level persistence assertion. Third, it evaluates the implemented boundary through 22 conformance and mutation tests and a fixed-output replay of three progressively stronger admission policies. The evaluation supports architectural feasibility for this path; it does not establish clinical readiness or population-level performance.

\section{Related work and security foundations}

\textit{Trustworthy health AI.} Governance and reporting frameworks require healthcare AI systems to make their data flows, limitations, human oversight, and lifecycle controls explicit~\cite{who2024,nist2024,futureai2025}. Akgun and Akgun further distinguish verification needs for factual retrieval, prediction, explanation, and recommendation~\cite{akgun2026}. These frameworks define assurance obligations at organizational or lifecycle level. The present work addresses a narrower implementation question: how an application can enforce one integrity decision between generated content and persistent record functions.

\textit{Grounding and fact verification.} Clinical information-extraction studies primarily measure whether models recover target fields from clinical text~\cite{agrawal2022,hein2025}. VeriFact instead evaluates whether claims in generated clinical text are supported by a patient's EHR through retrieval and an LLM judge~\cite{chung2025}. Verifiable summarization exposes source notes so clinicians can inspect generated summaries during chart review~\cite{verma2025}. Citation-quality research likewise shows that the presence of a reference does not guarantee that the reference supports the associated medical claim~\cite{wu2025}. These approaches improve support assessment or human verification, but support status does not by itself define which database writes and downstream uses are permitted.

\textit{Abstention and deterministic gates.} Selective prediction allows a model or auxiliary estimator to withhold an output when uncertainty or expected error cost is high~\cite{swaminathan2024}. Healthcare abstention research similarly treats answering and withholding as decisions under uncertainty and potential harm~\cite{presacan2026}. Both mechanisms focus on whether a model should answer. Our \emph{system-level abstention} occurs after the model answers: the surrounding application refuses promotion when independently checked evidence is insufficient. Deterministic integrity gates have also been applied to LLM-assisted manuscript production, where failed checks stop transitions between workflow stages~\cite{nam2026}. That work supports determinism where a mechanical check is sufficient. Our setting adds persistent patient records, purpose-specific reuse, and preservation of refused candidates.

\textit{Provenance and enforcement.} W3C PROV represents entities, activities, agents, and derivations~\cite{w3c2013}. HL7 FHIR Provenance applies related concepts to healthcare resources and records activities that create or update them~\cite{hl7provenance}. Provenance can support audit and reproducibility, but lineage recorded after a write does not establish that the write was authorized. The source--candidate relation must therefore participate in the admission decision.

Clark and Wilson distinguish unconstrained inputs from integrity-protected data and allow changes through certified transformation procedures~\cite{clarkwilson1987}. In our analogy, an LLM candidate is unconstrained input, a protected observation is constrained data, and evidence checking followed by projection is the transformation. The analogy explains why generation and admission are separate operations. We do not claim a full Clark--Wilson implementation because the prototype lacks organizational separation of duty, formal certification rules, and independent integrity-verification procedures.

The reference-monitor concept sharpens the enforcement requirement. A trustworthy monitor should mediate every protected operation, resist tampering, and remain small enough to analyze~\cite{anderson1972}. We treat the creation of an AI-derived observation as the protected operation. The prototype places a narrow monitor on the implemented laboratory write path, but it does not prove system-wide complete mediation or tamper resistance. The research gap is thus a stateful and enforceable transition: unsupported candidates must remain inspectable while being unable to enter the protected representation used by downstream record functions.

\begin{table}[t]
\centering
\caption{Positioning of evidence-gated promotion relative to adjacent controls.}
\label{tab:related-comparison}
\small
\begin{tabular}{@{}p{0.21\linewidth}p{0.25\linewidth}p{0.22\linewidth}p{0.23\linewidth}@{}}
\toprule
Control & Primary decision & Persistence control & Residual gap addressed here \\
\midrule
Extraction evaluation & Was a target field recovered? & None inherent & Recovered fields may still be unauthorized \\
Fact verification & Is a claim supported by retrieved evidence? & Usually external to the verifier & Support must govern the write path \\
Model abstention & Should the model answer? & None inherent & The application must refuse after an answer \\
Provenance & How was a resource produced? & Records a transition & Lineage alone does not authorize admission \\
Evidence-gated promotion & May this candidate be reused for purpose $p$? & Gate plus persistence assertion & Clinical truth and system-wide mediation remain open \\
\bottomrule
\end{tabular}
\end{table}

\subsection{Transition model}

Let $S$ be preserved source artifacts, $X$ bounded text packets derived from those artifacts, $C$ generated candidates, $R$ retained candidates that require review, and $P$ downstream purposes. A candidate $c\in C$ is a structured assertion with content fields, evidence fields, and generator-controlled metadata. A purpose $p\in P$ names an allowed reuse, such as inclusion in an observation history or export.

Source extraction is $\chi:S\rightarrow X$. For each purpose, $K_p:C\rightarrow\{0,1\}$ checks structural eligibility and $E_p:X\times C\rightarrow\{0,1\}$ checks source support independently of the generator. The admission predicate is
\begin{equation}
V_p(x,c)=K_p(c)\land E_p(x,c).
\end{equation}
For purpose-specific projection $\pi_p$, review mapping $\rho$, and lineage record $\ell$, routing is
\begin{equation}
T_p(s,x,c)=
\begin{cases}
\operatorname{admit}(\pi_p(c),\ell(s,x,c,p)), & x=\chi(s)\land V_p(x,c)=1,\\
\operatorname{review}(\rho(s,x,c)), & \text{otherwise.}
\end{cases}
\end{equation}
The output set $O_p$ contains the representations authorized for purpose $p$. Admission for one purpose does not imply admission for another: $\pi_p(c)\in O_p$ does not imply $\pi_{p'}(c)\in O_{p'}$ when $p\ne p'$. For example, evidence may justify showing ``glucose 90 mg/dL'' beside its source without justifying a clinical interpretation or alert. The current prototype implements one principal admission class for reuse as a document-derived observation; finer-grained policies remain part of the model rather than the evaluated implementation.

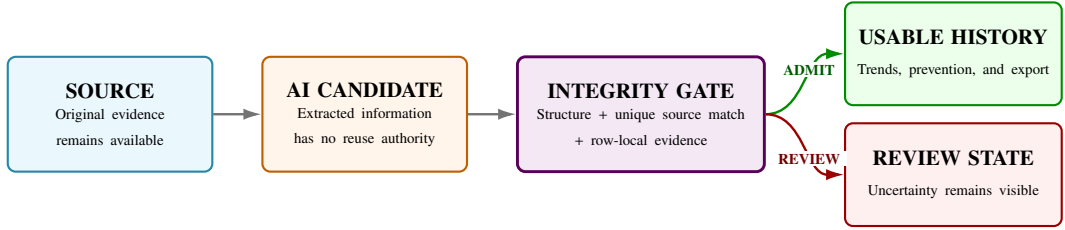
\begin{figure*}[]
\centering
\begin{tikzpicture}[scale=0.8, transform shape,
  box/.style={draw, rounded corners=3pt, align=center, minimum height=19mm, text width=31mm, inner xsep=4pt, inner ysep=5pt},
  source/.style={box, fill=cyan!7, draw=cyan!60!black, line width=0.8pt},
  candidate/.style={box, fill=orange!8, draw=orange!75!black, line width=0.8pt},
  gate/.style={box, fill=violet!9, draw=violet!70!black, line width=1.1pt, text width=38mm},
  accepted/.style={box, fill=green!8, draw=green!55!black, line width=0.8pt, text width=33mm, minimum height=17mm},
  review/.style={box, fill=red!6, draw=red!60!black, line width=0.8pt, text width=33mm, minimum height=17mm},
  flow/.style={-{Latex[length=2.4mm,width=1.5mm]}, line width=0.9pt, draw=black!55},
  flowlabel/.style={fill=white, inner xsep=3pt, inner ysep=1pt, font=\scriptsize\bfseries, text=black!65}
]
\node[source] (source) {\textbf{SOURCE}\\[2pt]{\scriptsize Original evidence\\remains available}};
\node[candidate, right=8mm of source] (candidate) {\textbf{AI CANDIDATE}\\[2pt]{\scriptsize Extracted information\\has no reuse authority}};
\node[gate, right=8mm of candidate] (gate) {\textbf{INTEGRITY GATE}\\[2pt]{\scriptsize Structure + unique source match\\+ row-local evidence}};
\node[accepted, right=13mm of gate, yshift=10mm] (record) {\textbf{USABLE HISTORY}\\[2pt]{\scriptsize Trends, prevention, and export}};
\node[review, right=13mm of gate, yshift=-10mm] (review) {\textbf{REVIEW STATE}\\[2pt]{\scriptsize Uncertainty remains visible}};
\draw[flow] (source) -- (candidate);
\draw[flow] (candidate) -- (gate);
\draw[flow, draw=green!55!black] (gate.east) to[out=0,in=180] node[flowlabel, pos=0.57, above, text=green!35!black]{ADMIT} (record.west);
\draw[flow, draw=red!60!black] (gate.east) to[out=0,in=180] node[flowlabel, pos=0.57, below, text=red!45!black]{REVIEW} (review.west);
\end{tikzpicture}
\caption{The generator proposes information, while a separate integrity gate grants or withholds purpose-limited reuse authority.}
\label{fig:model}
\end{figure*}

\subsection{Integrity properties}

The model defines five properties:
\begin{description}
\item[$P_1$ mediated promotion.] Every AI-derived observation in $O_p^{AI}$ has a passing monitored transition:
$o\in O_p^{AI}\Rightarrow\exists s,x,c:\ x=\chi(s)\land V_p(x,c)=1\land o=\pi_p(c)$.
\item[$P_2$, non-self-certification.] Let $m$ be generator-controlled metadata, including confidence or a claimed verification flag. For fixed source, quote, and asserted value, changing $m$ cannot change the authoritative decision:
$V_p(x,(q,v,m))=V_p(x,(q,v,m'))$. 
\item[$P_3$, provenance preservation.] Admission emits lineage $\ell(s,x,c,p)$ with the protected representation, so the source and policy decision remain inspectable. Full activity-level lineage additionally requires immutable candidate, verifier, and policy identifiers.

\item[$P_4$, fail-closed admission.] Evidence that is absent, unmatched, multiply matched, or inconsistent under policy routes to $R$, not $O_p$.

\item[$P_5$, uncertainty preservation.] Refusal does not erase the candidate or rewrite it as certainty. The review representation retains the proposed value and evidence context needed to inspect it.
\end{description}
The properties separate three assurance obligations. The policy must detect the inconsistencies within its scope, the enforcement path must prevent bypass, and the provenance layer must retain the basis of each decision. Passing $V_p$ establishes policy admissibility only.

\subsection{Threat model and conditional assurance}

The generator is potentially faulty or adversarial. It may return arbitrary schema-valid values, source identifiers, quotes, confidence scores, and positive verification claims. The protected asset is the integrity of AI-derived observations authorized for downstream reuse. Manual observations follow a separately identified user-input path and are outside $O_p^{AI}$. The security goal is that generator-controlled fields alone cannot authorize promotion.

The trusted computing base contains the code that derives source packets, validates evidence, routes candidates, projects observations, and performs the repository write, together with the database state on which those operations depend. Authentication and account isolation are assumed. Verification compares candidates with extracted text packets rather than raw document pixels. Parser or OCR corruption, document forgery, patient mis-association, malicious source content, compromised server code, and provider confidentiality therefore remain outside the integrity guarantee.

The central assurance claim is conditional. Assume that (A1) every write to $O_p^{AI}$ is mediated by $T_p$, (A2) the admission branch of $T_p$ executes only when $V_p(x,c)=1$, and (A3) trusted code computes $V_p$ independently of generator-controlled metadata. Under these assumptions, a candidate for which $V_p(x,c)=0$ cannot enter $O_p^{AI}$. If it did, A1 would imply an executed transition and A2 would require $V_p(x,c)=1$, contradicting the failed predicate. The prototype tests A2 and selected bypasses in one path. It does not prove A1 for the entire application or establish tamper resistance.

\section{Prototype instantiation}

Medical DataCloud is a full-stack personal medical-data workspace. Users can upload medical documents, inspect extracted information, review uncertain results, follow laboratory values over time, and export a structured history. The application is not an electronic health record, diagnostic system, or clinical decision-support system. It provides the persistent setting needed to study when AI-derived values become reusable record data.

The implemented laboratory path has six stages. First, the upload service validates a document, calculates its SHA-256 digest, encrypts the original file, and stores extracted page text with source and text hashes. Second, the ingestion service derives page and non-empty line packets. Each packet has a stable source identifier, page and optional line number, text, and checksum. Third, a hosted LLM returns a schema-constrained laboratory object containing candidate observations and evidence fields. Fourth, server code discards the authority of the returned \texttt{verified} field and recomputes evidence state against the packets.

Evidence checking first requires a non-empty, bounded quote with one source match. Exact matching records character offsets; normalized matching tolerates whitespace, punctuation, diacritics, spaced units, and decimal comma. Duplicate matches fail closed. For each numeric laboratory candidate, a second row-local check requires the analyte and numeric value on the same evidence line. The value must occur as a complete numeric token, so 9 does not match 90. If the candidate supplies a unit or numeric reference bounds, those fields must occur on the same line as well.

Fifth, the laboratory task maps every numeric candidate to a source-linked extracted fact. A supported candidate receives \texttt{verified}; a refused candidate receives \texttt{needs\_review}. Only verified candidates are projected into observation drafts. Sixth, the observation repository checks the verification flag again before any database mutation. The document, extracted facts, observations, summaries, and extraction metadata are persisted within the document-processing transaction. The second assertion is defense in depth for the tested repository interface, not a database constraint or proof that every possible write path is mediated.

\begin{table}[t]
\centering
\caption{Illustrative candidate-to-record transitions for one source row.}
\label{tab:end-to-end-example}
\small
\begin{tabular}{@{}p{0.20\linewidth}p{0.34\linewidth}p{0.37\linewidth}@{}}
\toprule
Stage & Supported candidate & Refused candidate \\
\midrule
Source row & \multicolumn{2}{l}{\texttt{Glucose 90 mg/dL 70--99}} \\
LLM value & \texttt{90 mg/dL} & \texttt{9 mg/dL}, \texttt{verified=true} \\
Independent check & Unique row contains analyte, token 90, unit, and bounds & Token 9 is absent; model verification is ignored \\
Stored fact & \texttt{verified} with source context & \texttt{needs\_review} with source context \\
Observation & Created for record reuse & Not created \\
\bottomrule
\end{tabular}
\end{table}

The implementation preserves the document identifier and hash, packet location, source identifier, quote, offsets, and extracted-fact validation state. Separate extraction-run records retain the input hash, model provider and name, prompt version, schema version, counts, and warnings. However, persisted facts and observations are not currently bound immutably to the exact extraction run, verifier version, and policy version that produced the decision. The prototype therefore implements source-level lineage but only partially satisfies $P_3$. It also lacks candidate correction followed by re-verification, revocation of admitted observations, and policy-versioned re-promotion.

\section{Evaluation}

\subsection{Evaluation questions and method}

The evaluation addresses three questions. First, does the implemented laboratory path conform to $P_1$, $P_2$, $P_4$, and $P_5$ under selected faults and bypass attempts? Second, how does evidence-policy strength change automatic admission and review routing when generator output is held fixed? Third, how well do the emitted candidates agree with manually labelled source rows on fields that the gate does and does not check?

Three developer-written suites contain 22 conformance and mutation cases: 17 evidence-validation tests, three laboratory-task tests, and two repository-boundary tests. The cases cover exact and normalized matches; missing, wrong-source, duplicate, and oversized evidence; analyte, value, unit, and range disagreement; numeric substrings; neighboring-row borrowing; a forged positive verification flag; verified-only projection; and an unverified persistence attempt. These tests exercise observable interfaces in one application path. They are not formal verification or independent penetration testing.

The policy replay uses saved outputs from nine laboratory PDFs with 102 manually labelled rows. The labels are representative row subsets derived from source transcripts, not complete-document abstractions, and were not adjudicated by clinicians or a second rater. The outputs were generated through OpenRouter with \texttt{openai/gpt-4o-mini}, temperature zero, and prompt version \texttt{medical-extraction:v1}. Reusing the same 97 numeric candidates removes model-run variation from the policy comparison.

We compare three policies. \emph{Schema only} admits every numeric candidate that passed the output schema. The \emph{packet-level} policy uses the evidence decisions saved during the original extraction run, which required source-text support but did not enforce the current row-local field checks. The \emph{row-local} policy replays each candidate through the current verifier, requiring a unique source anchor and same-row support for the analyte, numeric value, supplied unit, and supplied bounds. Failure means that the policy did not establish support; it does not establish that the medical value is false.

The evaluation used historical laboratory reports belonging to one adult member of the research team, who authorized their use for this study. No other data subjects were included. The institutional research ethics body determined that formal approval was not required for this software evaluation. Only aggregate results are reported. Raw reports, extracted transcripts, and per-document model outputs contain personal health information and are not publicly available. Report content was processed through OpenRouter using an OpenAI model; third-party handling was governed by the API policies applicable during execution, and no zero-retention claim is made. Private artifacts are retained only for research verification and will be deleted when no longer required for the study; the data subject controls earlier deletion. They are excluded from public data availability, and no PHI-containing repository is provided as a research artifact.

\subsection{Results}
\begin{table}
\centering
\caption{Policy replay on the same 97 numeric candidates.}
\label{tab:results}
\small
\begin{tabular}{@{}lrrr@{}}
\toprule
Admission policy & Admitted & Review & Admission rate \\
\midrule
Schema only & 97 & 0 & 100.0\% \\
Saved packet-level check & 94 & 3 & 96.9\% \\
Hardened quote/row-local gate & 72 & 25 & 74.2\% \\
\bottomrule
\end{tabular}
\end{table}

All 22 tests passed. Server code recomputed the forged verification claim, ambiguous and cross-row evidence failed closed, verified-only mapping excluded unsupported candidates, and the repository rejected an unverified write before mutation. The tests support conformance of the exercised interfaces, not complete mediation across the whole application.

\begin{table}[H]
\centering
\caption{Conformance evidence for the implemented trust boundary.}
\label{tab:conformance}
\small
\begin{tabular}{@{}p{0.22\linewidth}p{0.10\linewidth}p{0.59\linewidth}@{}}
\toprule
Threat exercised & Property & Enforced outcome \\
\midrule
Self-certification & $P_2$ & A forged positive verification flag is ignored and the evidence decision is recomputed. \\
Unsupported or ambiguous evidence & $P_4$ & Duplicate, cross-row, substring, absent, and field-inconsistent support routes to review. \\
Projection or persistence bypass & $P_1$ & Verified-only mapping excludes the candidate, and the repository rejects an unverified write before mutation. \\
Loss of uncertainty on refusal & $P_5$ & The candidate and its evidence context remain available as a review item. \\
\bottomrule
\end{tabular}
\end{table}

The extractor matched 80 of 102 labelled rows (78.4\% labelled-row coverage) and emitted 97 numeric candidates. Schema-only admission accepted all 97 candidates. The packet-level decision accepted 94, while the row-local gate accepted 72 and retained 25 for review. Stronger locality therefore reduced automatic admission by 25.8 percentage points relative to schema-only routing. This is an authority-workload trade-off, not a measured reduction in clinical error.

The denominators answer different questions. Labelled-row coverage uses 102 source rows: 80 were matched and 22 were not. Admission uses all 97 emitted numeric candidates: 80 could be paired with a label and 17 could not. Because annotation covered representative row subsets rather than every document row, those 17 candidates cannot be classified as false positives.

On the 80 matched rows, agreement with labels was 100\% for units (80/80), 94.6\% for lower bounds (70/74), 93.8\% for upper bounds (75/80), 86.3\% for interpretation (69/80), and 85.0\% for effective dates (68/80). The gate checks the local occurrence of selected fields, but it does not validate the semantics of effective dates or interpretation codes. These results show why source support and semantic correctness require separate evaluations.

\section{Discussion}

The main contribution is a change in the location of authority. Prompting constrains generator behavior, confidence estimates uncertainty, fact verification assesses support, and human review asks a person to detect problems. Evidence-gated promotion complements these controls by assigning the database transition to a separate deterministic component. The conformance tests support a conditional containment claim for the evaluated path: a failed candidate is retained as a fact but does not become a document-derived observation. This claim would be refuted by an unmediated write path or a gate decision controlled by model metadata.

The replay also shows that groundedness depends on the evidence predicate. Packet-level co-occurrence admitted candidates that the row-local policy refused. The stronger policy is designed to resist duplicate evidence, cross-row borrowing, and numeric-substring matches, but the replay does not establish that every refusal was correct. Its clearest empirical effect is an increase in review routing from 3.1\% to 25.8\%. A useful deployment study must therefore measure false admission, false review, correction outcomes, and reviewer time instead of treating a lower admission rate as automatically safer.

The deterministic gate does not require an additional model call. Its work consists of bounded quote search, normalization, and same-row field matching, so its cost grows with the number of candidates and the amount of packet text searched. The present replay did not isolate gate latency or memory use. Computational overhead is therefore expected to be small relative to hosted generation but remains unmeasured. Human overhead is potentially more important: Medical DataCloud exposes refused facts and their source context in the document view, but it does not yet provide a complete correct-and-reverify workflow or a clinician-facing review interface.

An EHR deployment could place the gate between document ingestion and creation of a FHIR Observation. Refused candidates would remain in a quarantined review representation; admitted candidates could become Observation resources accompanied by Provenance records that identify the source document, extraction activity, verifier, and policy version~\cite{hl7provenance}. Such integration would require every create, update, import, and bulk-write path to use the same admission service. Database constraints, service credentials, audit events, revocation, and version-specific Provenance links would be needed before claiming complete mediation in a clinical system.

The model can extend beyond laboratories only through domain-specific predicates. Medication, allergy, diagnosis, and narrative candidates require different identity, status, negation, and temporal checks. The present results cannot be generalized to those domains or across patients, laboratories, languages, scanners, and report layouts.

Integrity admission does not authorize disclosure or indefinite retention. Because sources, quotes, refused candidates, and lineage increase the sensitive material held for audit, deployments need separate controls for minimization, access, third-party processing, export, and deletion~\cite{who2024,nist2024}.

The lexical verifier cannot establish the authenticity of the document, the identity of the patient, the correctness of the OCR, the clinical meaning or the medical correctness. The subset labels lack clinician adjudication and inter-rater agreement, while the tests lack independent security assessment. A stronger study should use consented multi-subject and multi-center reports, complete double annotation, predeclared false-admission and false-review metrics, timed review tasks, adversarial documents, penetration testing and prospective workflow evaluation following guidelines such as DECIDE-AI~\cite{decideai2022}.

\section{Conclusion}

Generated health information becomes more consequential when a persistent system reuses it. Evidence-gated trust promotion separates generation from that authorization decision. In the Medical DataCloud laboratory path, deterministic source checks, verified-only mapping, and a repository assertion prevent selected unsupported candidates from entering the observation history while preserving them for review. Twenty-two tests demonstrate conformance of the exercised interfaces, and the fixed-output replay shows that stronger source locality reduces automatic admission while increasing review demand. The study establishes the feasibility for a narrow integrity boundary. It does not establish clinical truth, clinical safety, acceptable workflow burden, or complete mediation across a healthcare system.

\section*{Code availability}
The code is available at \url{https://github.com/noragirda/medicloud}. The online demo is available at \url{https://medicloud-theta.vercel.app/}

\section*{Acknowledgment} 
A. Groza is supported by a grant of the Ministry of Research, Innovation and Digitization, CCCDI-UEFISCDI, project number
PN-IV-P6-6.3-SOL-2024-2-0312 within PNCDI IV.

\end{document}